# DETECTING TEXT-LEVEL INTELLECTUAL INFLUENCE WITH KNOWLEDGE GRAPH EMBEDDINGS

*Lucian Li, Eryclis Silva,*

*School of Information Sciences, University of Illinois, Urbana-Champaign*

## Abstract

**Introduction**

Tracing the spread of ideas and the presence of influence is a question of special importance across a wide range of disciplines, ranging from intellectual history to cultural analytics, computational social science, and the science of science.

**Method**

We collect a corpus of open source journal articles, generate Knowledge Graph representations using the Gemini LLM, and attempt to predict the existence of citations between sampled pairs of articles using previously published methods and a novel Graph Neural Network based embedding model.

**Results**

We demonstrate that our knowledge graph embedding method is superior at distinguishing pairs of articles with and without citation. Once trained, it runs efficiently and can be fine-tuned on specific corpora to suit individual researcher needs.

**Conclusion(s)**

This experiment demonstrates that the relationships encoded in a knowledge graph, especially the types of concepts brought together by specific relations can encode information capable of revealing intellectual influence. This suggests that further work in analyzing document level knowledge graphs to understand latent structures could provide valuable insights.

# Introduction

Tracing the spread of ideas and the presence of influence is a question of special importance across a wide range of disciplines, ranging from intellectual history to cultural analytics, computational social science, and the science of science. While some fields (such as bibliometrics) generally rely on annotated citation datasets, such data is limited to past few decades and the domain of scholarly publication. A wide range of potential questions rely on solely on unstructured text.

Previous attempts to explore and discover examples of influence in corpora have not been previously validated on annotated corpora, so we have no sense of the false negative rate. We propose a novel method to calculate pairwise influence between documents and evaluate it against previously published methods.

We generate individual knowledge graphs for each document and then calculate Knowledge Graph embeddings. We experiment with node level embeddings like TransE (Bordes et al, 2013) but find that the computational costs of generating embeddings using the combined knowledge graphs across our article corpus proved prohibitive. Other approaches, such as anchoring, proved infeasible because of the relatively small size of individual knowledge graphs.

Our final approach utilizes Graph Convolutional Networks to train an embedding model through the contrastive task of identifying pairs of knowledge graphs generated using the Gemini LLM. The intuition is that each knowledge graph represents a conceptual space emerging from arguments and claims advanced in the document and that structural and semantic similarities in knowledge graphs correspond to latent structural similarities in the text and argumentative style.

# Related Work

## Influence Detection

With the large-scale digitization of archives, the past decade has seen a range of different approaches to detect influence and similarity across documents. These have relied on three families of methods broadly: text reuse, topic models, and word sense similarity.

Text reuse methods, such as (Funk and Mullen, 2018) and (Cordell et al., 2024), essentially use n-gram overlap to detect large runs of text shared between documents. While able to detect all instances of direct quotation, these methods cannot detect paraphrase or indirect influence.

Next, several studies, such as (Rockmore et al., 2018) and (Barron et al., 2018) have compared topic model distributions across documents. While these methods capture high level changes in subjects and disciplinary focus, they are unable to evaluate specific changes in argumentation or offer any insight into intersections in semantic content.

Finally, another family of recent studies focus on detecting similarities in word sense, either by calculating similarities in word embeddings for the same word across different documents as in (Soni et al., (21) or calculating the BERT perplexity for unexpected words as in (Vicinanza et al. , 2023). These methods focus on influence in the domain of language use and word choice. They can capture stylistic influence and innovative language, but again fail to capture semantic or argumentative similarity.

Lastly, some studies have leveraged knowledge graphs to calculate conceptual similarity, although these methods use knowledge graphs to encode information rather than calculating similarity in knowledge graph structure (Zhang and Zhu, 2022) (Bi et al., 2022).

### Knowledge Graph Generation

Previous triplet extraction tools include traditional NLP based entity and relation extraction pipelines such as Kim et al. (2021) and LLM based approaches like Melnyk et al. (2021), Bronzini et al. (2023), Sun et al. (2024), and Han et al. (2023). We eventually used Google's Gemini Team et al. (2023) LLM because of its relatively high accuracy, free access, and relatively low implementation complexity.

### Knowledge Graph Embeddings

A range of projects, such as Li et al. (2023), have worked on using knowledge graph embeddings to calculate similarity. These, however, generally focus on creating large single knowledge graphs from documents, citations, and metadata. Our method, which aims to align multiple independent knowledge graphs, is closer to approaches such as Fanourakis et al. (2023), which compares embeddings across different knowledge graphs to align entities. Although their goal is to combine different KG, this shows that embeddings across independent KG may be comparable. Baumgartner et al. (2023) and Huang et al. (2022) also attempt to align entities across different KG, but they find that base embeddings may need additional transformations or anchoring to be effectively compared across different embedding spaces. Our final approach also draws from previous attempts to create a Graph Convolutional Network based embedding system that encodes graph structure through contrastive learning, applied in knowledge graphs by Xu et al. (2019). We apply a generally similar training architecture, but incorporate a different strategy to select contrastive pairs and incorporate additional semantic data in node features

### Knowledge Graph Evaluation Metrics

Knowledge Graph (KG) evaluation is a critical area of research that has seen various approaches proposed in recent years. These approaches can be broadly categorized into structure-based, data quality-based, and task-oriented evaluation methods.

Structure-based evaluation metrics have been a primary focus of several studies. Seo et al. (2022) presented novel metrics such as the Instantiated Class Ratio, Instantiated Property Ratio, Class Instantiation, Subclass Property Acquisition, and Subclass Property Instantiation. They also proposed the Inverse Multiple Inheritance metric to assess ontology complexity. These metrics provide valuable insights into the structural quality of knowledge graphs.

Complementing structure-based approaches, Dou et al. (2023) proposed different measures of knowledge in KGs, introducing more broadly applicable metrics. Their K Score, I Score, and C Score, derived from the science of science, information theory, and causality perspectives respectively, offer a multifaceted approach to KG evaluation. This work aims to address limitations in existing structure-based and data quality-based assessment techniques.

Moving beyond intrinsic evaluation methods, Heist et al. (2023) presented a framework to evaluate KGs via downstream tasks. This framework enables a comprehensive assessment of KGs by evaluating their performance on multiple kinds of tasks such as classification, regression, and recommendation. Their experiments revealed significant variations in KG performance depending on the specific task, highlighting the importance of extrinsic evaluation metrics as a complement to established intrinsic criteria.

These diverse approaches demonstrate the complexity of KG evaluation and the need for multifaceted assessment methods that consider both intrinsic quality and practical utility in downstream applications.

## Method

### Dataset

As an evaluation dataset, we used the Semantic Scholar API to download pairs of recent academic articles on specific subjects (Darwinism, terrorism, convolutional graph learning, carbon offset, and MECP2, a neurological disease gene). These subjects were selected to cover a wide range of disciplines to evaluate the ability of the method to generalize across disciplines instead of fitting to norms of one area.

We downloaded the text of roughly 4,400 open-source articles divided evenly into about 900 per subject.

From these articles we sampled pairs. We generate roughly 22,000 total pairs, of which 8,500 are "positive samples" where one article cites the other. We also generate 13,500 negative samples, which are pairs of articles about the same subject that do not cite each other. This was done to prevent the model from learning only to distinguish article topics.

### Knowledge Graph Generation

We initially explored using the Mistral 7B model locally for our knowledge graph generation task. However, through subjective evaluation, we found that Gemini Pro consistently produced more meaningful and relevant entity-relationship triples compared to Mistral 7B.

Given the dataset at hand, our approach with Gemini Pro entailed systematically sampling one-third of the dataset to assess the model's efficacy in extracting entities and relationships. We began by partitioning the textual data, carefully considering the context window constraints to optimize model performance. Utilizing LangChain we employed recursive character segmentation, fine-tuning parameters such as size and overlap.

After meticulously customizing prompts for each text chunk, we applied the Gemini Pro model. Challenges arose during entity extraction due to the presence of numerous citations, which prompted us to refine the prompt, instructing the model to disregard direct citations. As a result, we obtained JSON lists containing two nodes and one edge each, which were then utilized to enhance the original data frame.

In our approach with Gemini Pro, we adopted a one-shot learning paradigm to train our model, a machine learning technique where a model learns to recognize patterns or make predictions based on only a single example or a few examples of each class. By leveraging one-shot learning, our model can effectively extract entities and relationships from the dataset with minimal labelled examples, enhancing efficiency and reducing the need for extensive data annotation.

**Figure 1.** sample knowledge graph generated from Gemini

**Figure 2.** Methods flowchart

## Analysis

### Graph Neural Network embedding

Using our positive and negative sampled pairs, we conducted contrastive training using a three layer Graph Convolutional Network (GCN).

For the node features, we generate semantic embeddings using GTE Li et al. [2023], a BERT based sentence embedding method. These 384-dimension semantic representations of the node content are used as node features. In theory, including these representations as node attributes will allow the model to treat nodes conveying similar concepts in the same way.

We initialize 500 dimensions for the hidden layer, and 100 dimensions for the final graph embedding. the 100-dimensional final layer representations were mean pooled into one 100-dimensional graph embedding. Loss was calculated with the pyTorch CosineEmbeddingLoss function, which penalizes high distance between true pairs and low distance between false pairs. Loss was backpropagated and optimized with the Adam optimizer.

Currently, hyperparameter tuning through grid search is infeasible because of computational demands. We use the DGL implementation of GCN with a pyTorch backend.

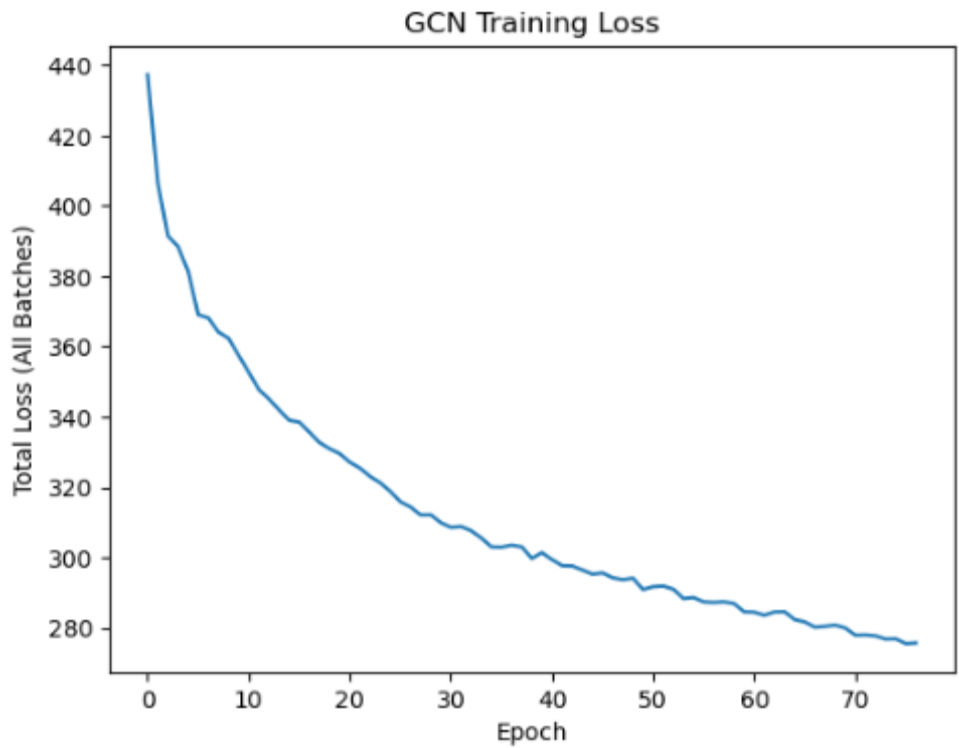

**Figure 3.** Training error in GCN training

We observe near convergence after 80 epochs of training. Other hyperparameters are contrastive margin=0.5, learning rate=0.05 and an 80% train-test split.

We also experimented with a Graph Attention Network to weight different types of relationships differently, but the computational demands proved too high for our limited resources.

The final model is able to take a knowledge graph as input and output a 100 dimensional embedding encoding argumentative structure.

### Reproduction of previously published approaches

We implement three other methods for a performance comparison. For text reuse detection, we use the text-matcher package Reeve [2020]. We use default settings and remove the first 300 and last 2000 characters to remove titles and listed citations to limit the comparison to the argument text. Text reuse pairs were scored by raw number of overlapping segments. The topic model was implemented with Gensim. After some hyperparameter tuning, we used LDA with 500 topics as that produced the best results. LDA vectors were scored using KL divergence. Finally, GTE embeddings were implemented with the HuggingFace sentence-transformers package. We created 1000 characters chunks, which were embedded and then summed to generate a document embedding. GTE vector pairs were evaluated using cosine distance.

## Results

### Evaluation metrics:

Because there is no gold standard way to evaluate influence, we used the presence of citation as a proxy metric. The overall goal for evaluation is the citation prediction task. Given two articles about the same topic, we will evaluate the ability of models to predict if a citation exists between the two articles. We report three performance metrics: we compare the distribution of scores reported by each method for true and false pairs. We use a rank-sum test to evaluate the likelihood that the scores for true pairs are significantly higher than those for false pairs. We also calculate the Receiver Operating Characteristic curve for each method. This shows potential trade-offs in performance when each method is used as a classifier to distinguish true and false pairs. Finally, we calculate the optimal threshold for each method using the ROC curve and evaluate the F1-score performance of each method. These metrics capture the distribution of scores produced as well as their potential as a classification model.

### Performance

We observe superior performance for our model on all metrics. While all previously published models had significant ability to discriminate true pairs than false pairs, our model achieved high significance of separation for the rank sum metric. This suggests that the separation between the scores for true and false pairs is superior using the KG embeddings.

| Metric | Text Reuse | Topic Model KL Divergence | Document Embedding (GTE) distance | KG Embedding (Ours) |
|---|---|---|---|---|
| Rank sum test (inverse log p-value) | 75 | 191 | 267 | **611** |
| Area under ROC curve | 0.55 | 0.58 | 0.59 | **0.61** |
| F1 score (optimal threshold) | 0.17 | 0.46 | 0.48 | **0.55** |

**Table 1.** accuracy metric comparison

Our model has a higher area under the ROC, and furthermore dominates the previous models at all points of the curve. At the optimal threshold, our method achieves a higher F1 accuracy score than any published method. Most prior methods achieve sub 0.5 F1-score, suggesting that this is inherently a very difficult problem.

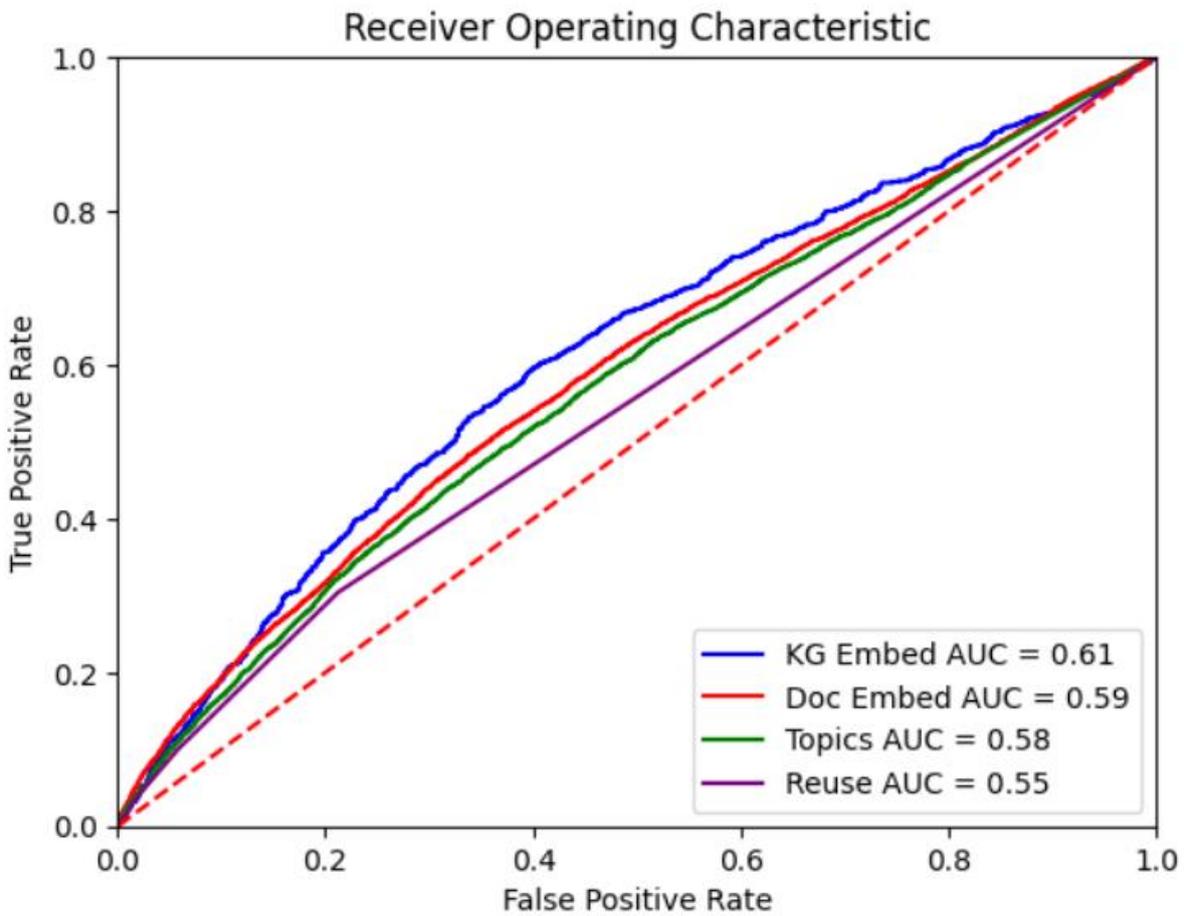

**Figure 4.** ROC curves for different methods. Our method in blue is superior across all accuracy thresholds.

The text reuse method specifically performed worse because of the very high number of both true and false documents with no exact text reuse present at all. Embedding based methods, like sentence embeddings and KG embeddings can capture a wider range of similarity with greater granularity.

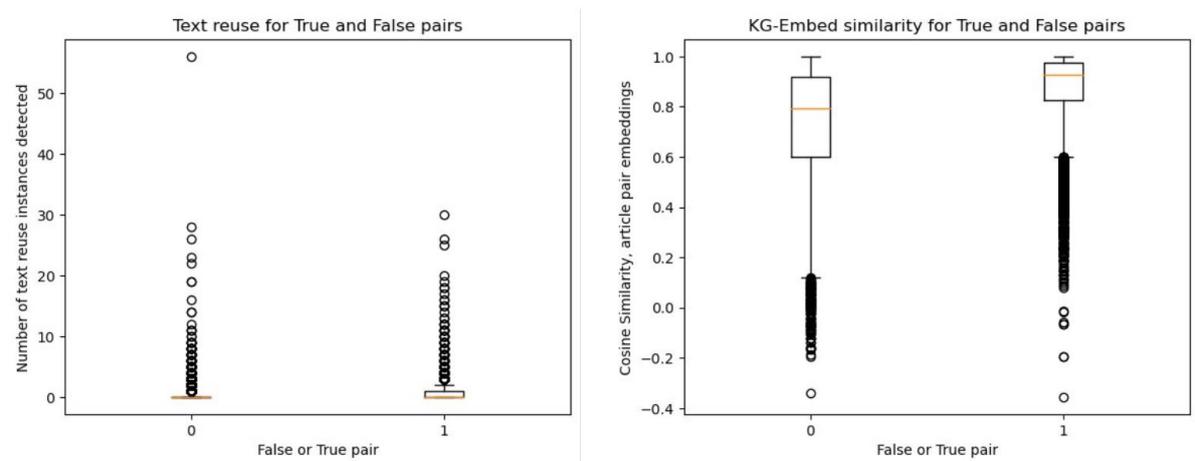

**Figure 5.** text reuse vs embedding similarity across documents

## Conclusion

We report a knowledge graph-based method for detecting influence that outperforms previously published methods. Our method is trained through a contrastive approach to distinguish citation pairs from non-citation pairs of similar articles. This approach incorporates graph/argumentative structure and the semantic information encoded in knowledge graphs.

One major issue we faced was lack of access to GPU acceleration and compute credits. More complex architectures like Graph Attention Networks, more neural network layers, additional epochs of training, or additional training data would have produced a more effective model. Additional hyperparameter tuning could have produced significantly improved results as well. For future research, additional close reading of successful and unsuccessful cases is necessary. While our method outperforms on quantitative metrics, it is important to understand if it is detecting types of influence previously missed. Potentially, an ensemble method of our KG embedding approach, topic models, and sentence embeddings can achieve even better results by covering for weaknesses for specific models and capturing a wider range of evidence of influence. We also plan to evaluate the method on out of domain data, especially to establish the potential transfer learning ability of the model to generalize to patterns in different contexts.

We have demonstrated that the relationships between nodes and edges in knowledge graphs encode useful information that a neural network can learn to detect latent similarities between documents. We can think of a knowledge graph as representing the conceptual space of a document-our project shows that studying the structures and relationships of this space can provide valuable insights about relationships between documents.

## References

Alexander TJ Barron, Jenny Huang, Rebecca L Spang, and Simon DeDeo. Individuals, institutions, and innovation in the debates of the French revolution. Proceedings of the National Academy of Sciences, 115(18):4607–4612, 2018.

Matthias Baumgartner, Daniele Dell'Aglio, Heiko Paulheim, and Abraham Bernstein. Towards the web of embeddings: Integrating multiple knowledge graph embedding spaces with fedcoder. Journal of Web Semantics, 75:100741, 2023.

Sheng Bi, Zafar Ali, Meng Wang, Tianxing Wu, and Guilin Qi. Learning heterogeneous graph embedding for Chinese legal document similarity. Knowledge-Based Systems, 250:109046, 2022. ISSN 09507051. doi:https://doi.org/10.1016/j.knosys.2022.109046. URL https://www.sciencedirect.com/science/article/pii/S0950705122005123.

Antoine Bordes, Nicolas Usunier, Alberto Garcia-Duran, Jason Weston, and Oksana Yakhnenko. Translating embeddings for modelling multi-relational data. Advances in neural information processing systems, 26, 2013.

Marco Bronzini, Carlo Nicolini, Bruno Lepri, Andrea Passerini, and Jacopo Staiano. Glitter or gold? deriving structured insights from sustainability reports via large language models. arXiv preprint arXiv:2310.05628, 2023.

Ryan Cordell, David A Smith, Abby Mullen, and Jonathan D Fitzgerald. Going the Rounds Virality in Nineteenth Century American Newspapers. University of Minnesota Press, Forthcoming, 2024.

Jacob Devlin, Ming-Wei Chang, Kenton Lee, and Kristina Toutanova. Bert: Pre-training of deep bidirectional transformers for language understanding. arXiv preprint arXiv:1810.04805, 2018.

Dou, Jason Xiaotian, Haiyi Mao, Runxue Bao, Paul Pu Liang, Xiaoqing Tan, Shiyi Zhang, Minxue Jia, Pengfei Zhou, and Zhi-Hong Mao. "The Measurement of Knowledge in Knowledge Graphs." In Proceedings of the AAAI 2023 Workshop on Representation Learning for Responsible Human-Centric AI (R2HCAI). Washington, DC, USA: Association for the Advancement of Artificial Intelligence (AAAI), 2023.

Nikolaos Fanourakis, Vasilis Efthymiou, Dimitris Kotzinos, and Vassilis Christophides. Knowledge graph embedding methods for entity alignment: experimental review. Data Mining and Knowledge Discovery, 37(5):2070–2137, 2023.

Kellen Funk and Lincoln A Mullen. The spine of American law: Digital text analysis and us legal practice. The American Historical Review, 123(1):132–164, 2018.

Jiuzhou Han, Nigel Collier, Wray Buntine, and Ehsan Shareghi. Pive: Prompting with iterative verification improving graph-based generative capability of llms. arXiv preprint arXiv:2305.12392, 2023.

Nicolas Heist, Sven Hertling, and Heiko Paulheim. 2023. KGrEaT: A Framework to Evaluate Knowledge Graphs via Downstream Tasks. In Proceedings of the 32nd ACM International Conference on Information and Knowledge Management (CIKM '23). Association for Computing Machinery, New York, NY, USA, 3938–3942. https://doi.org/10.1145/3583780.3615241.

Hongren Huang, Chen Li, Xutan Peng, Lifang He, Shu Guo, Hao Peng, Lihong Wang, and Jianxin Li. Cross-knowledge graph entity alignment via relation prediction. Knowledge-Based Systems, 240:107813, 2022. ISSN 09507051. doi: https://doi.org/10.1016/j.knosys.2021.107813

Vlad-Iulian Ilie, Ciprian-Octavian Truică, Elena-Simona Apostol, and Adrian Paschke. Context-aware misinformation detection: A benchmark of deep learning architectures using word embeddings. IEEE Access, 9:162122–162146,2 021. doi:10.1109/ACCESS.2021.3132502.

Taejin Kim, Yeoil Yun, and Namgyu Kim. Deep learning-based knowledge graph generation for covid-19. Sustainability, 13(4):2276, 2021.

Guangtong Li, L Siddharth, and Jianxi Luo. Embedding knowledge graph of patent metadata to measure knowledge proximity. Journal of the Association for Information Science and Technology, 74(4):476–490, 2023.

Igor Melnyk, Pierre Dognin, and Payel Das. Grapher: Multi-stage knowledge graph construction using pretrained language models. In NeurIPS 2021 Workshop on Deep Generative Models and Downstream Applications, 2021.

Jonathan Reeve. Text-matcher. https://github.com/JonathanReeve/text-matcher, 2020.

Daniel N Rockmore, Chen Fang, Nicholas J Foti, Tom Ginsburg, and David C Krakauer. The cultural evolution of national constitutions. Journal of the Association for Information Science and Technology, 69(3):483–494, 2018.

Seo, Sumin, Heeseon Cheon, Hyunho Kim and Dongseok Hyun. Structural Quality Metrics to Evaluate Knowledge Graphs. ArXiv preprint arXiv:2211.10011, 2022.

Sandeep Soni, Lauren F Klein, and Jacob Eisenstein. Abolitionist networks: Modeling language change in nineteenth century activist newspapers. Journal of Cultural Analytics, 6(1), 2021.

Qi Sun, Kun Huang, Xiaocui Yang, Rong Tong, Kun Zhang, and Soujanya Poria. Consistency guided knowledge retrieval and denoising in llms for zero-shot document-level relation triplet extraction, 2024.

Gemini Team, Rohan Anil, Sebastian Borgeaud, Yonghui Wu, Jean-Baptiste Alayrac, Jiahui Yu, Radu Soricut, Johan Schalkwyk, Andrew M Dai, Anja Hauth, et al. Gemini: a family of highly capable multimodal models. ArXiv preprint arXiv:2312.11805, 2023.

Paul Vicinanza, Amir Goldberg, and Sameer B Srivastava. A deep-learning model of prescient ideas demonstrates that they emerge from the periphery. PNAS nexus, 2(1):pgac275, 2023.

Kun Xu, Liwei Wang, Mo Yu, Yansong Feng, Yan Song, Zhiguo Wang, and Dong Yu. Cross-lingual knowledge graph alignment via graph matching neural network. arXiv preprint arXiv:1905.11605, 2019.

Jinzhu Zhang and Lipeng Zhu. Citation recommendation using semantic representation of cited papers' relations and content. Expert Systems with Applications, 187:115826, 2022. ISSN 0957-4174. doi: https://doi.org/10.1016/j.eswa.2021.115826